\documentclass[onecolumn,pre,superscriptaddress,notitlepage]{revtex4-1}

\usepackage{hyperref}
\usepackage{url}

\usepackage{amssymb}
\usepackage{amsmath}
\usepackage{epsfig}
\usepackage{subfigure}
\usepackage{graphicx}
\usepackage{textcomp}
\usepackage{float}
\usepackage[normalem]{ulem}
\usepackage{color}
\usepackage{epstopdf}
\usepackage{subfigure}

\newcommand{\cin}{c_\mathrm{in}}
\newcommand{\cout}{c_\mathrm{out}}
\newcommand{\tc}{\sigma}
\newcommand{\bc}{c}

\newcommand{\id}{\mathbb{I}}
\newcommand{\e}{\mathrm{e}}
\newcommand{\Poi}{\mathrm{Poi}}
\newcommand{\Aspatial}{\mathbb{A}^{\mathrm{spatial}}}
\newcommand{\Dspatial}{\mathbb{D}^{\mathrm{spatial}}}
\newcommand{\Atemp}{\mathbb{A}^{\mathrm{temp}}}
\newcommand{\Dtemp}{\mathbb{D}^{\mathrm{temp}}}

\begin{document}
\title{Detectability thresholds and optimal algorithms \\ for community structure in dynamic networks}

\author{Amir Ghasemian}
\email[]{amir.ghasemianlangroodi@colorado.edu}
\affiliation{Department of Computer Science, University of Colorado, Boulder, CO 80309, USA}

\author{Pan Zhang}
\email[]{pan@santafe.edu}
\affiliation{Santa Fe Institute, 1399 Hyde Park Road, Santa Fe, NM 87501, USA}

\author{Aaron Clauset}
\email[]{aaron.clauset@colorado.edu}
\affiliation{Department of Computer Science, University of Colorado, Boulder, CO 80309, USA}
\affiliation{BioFrontiers Institute, University of Colorado, Boulder, CO 80305, USA}
\affiliation{Santa Fe Institute, 1399 Hyde Park Road, Santa Fe, NM 87501, USA}

\author{Cristopher Moore}
\email[]{moore@santafe.edu}
\affiliation{Santa Fe Institute, 1399 Hyde Park Road, Santa Fe, NM 87501, USA}

\author{Leto Peel}
\email[]{leto.peel@colorado.edu}
\affiliation{Department of Computer Science, University of Colorado, Boulder, CO 80309, USA}

\begin{abstract}
We study the fundamental limits on learning latent community structure in dynamic networks.  Specifically, we study dynamic stochastic block models where nodes change their community membership over time, but where edges are generated independently at each time step.  In this setting (which is a special case of several existing models), we are able to derive the detectability threshold exactly, as a function of the rate of change and the strength of the communities.  Below this threshold, we claim that no algorithm can identify the communities better than chance.
We then give two algorithms that are optimal in the sense that they succeed all the way down to this limit. The first uses belief propagation (BP), which gives asymptotically optimal accuracy, and the second is a fast spectral clustering algorithm, based on linearizing the BP equations. We verify our analytic and algorithmic results via numerical simulation, and close with a brief discussion of extensions and open questions.
\end{abstract}

\maketitle

%\section{Introduction}
%\label{Intro_section}
Relational or interaction variables are common feature of modern data sets, and these are often represented as a network. Examples include friendships or communication within a social network, regulatory interactions among genes, transportation between cities, and relations or hyperlinks in information systems. Many, perhaps most of these systems are also dynamic in nature, and their evolving structure is commonly represented as a sequence of graphs~\cite{clauset2007persistence,berger2010dynamic,gauvin2014detecting,kim2013nonparametric,mucha2010community, rossi2011modeling,xing2010state,zhu2014scalable}. 
Recently, a variety of techniques have been developed for automatically detecting communities---a task that is similar to traditional clustering~\cite{von2012clustering}, but on graphs---in these dynamic networks. These techniques include variants of multilayer or temporal modularity optimization~\cite{mucha2010community, bassett2013robust, bazzi2014community}, non-negative matrix or tensor factorization~\cite{acar2009link, dunlavy2011temporal, gauvin2014detecting, rossi2011modeling,zhu2014scalable}, minimum description length~\cite{sun2007graphscope,rosvall2010mapping}, and probabilistic models~\cite{yang2009bayesian,xing2010state,kim2013nonparametric,xu2014dynamic,han2014consistent,peixoto2015inferring,valles2014multilayer}. See Refs.~\cite{aggarwal2014evolutionary, hartmann2014clustering} for reviews. Despite these advances, relatively little is known about their optimality or the fundamental difficulty of detecting community structure in dynamic networks. In this paper, we derive a mathematically precise threshold on the detectability of communities in dynamic networks and give two algorithms that are optimal in the sense that they succeed all the way down to this  threshold.

Community detection in dynamic networks inherits many of the challenges of community detection in static networks, including learning the number of communities, their sizes and node membership, and the pattern of connections among communities, e.g., assortative, disassortative, core-periphery, etc. It also poses new challenges, because both the network edges and the community memberships may evolve over time. A common approach is to simply take the union of dynamic graphs over a certain time window, and treat the resulting graph with techniques from static network analysis~\cite{clauset2007persistence}, thereby ignoring the dynamics within the window. Here, we explicitly model the dynamic nature of these networks and the way community memberships change over time,  integrating information about the communities in an optimal way.

Our approach relies on probabilistic generative models, which can be used to learn latent community structure in real networks via Bayesian inference and to generate synthetic networks with known structure that can be used as benchmarks. A number of such models have recently been proposed for detecting communities in dynamic networks~\cite{kim2013nonparametric,xu2014dynamic}, including those based on the stochastic block model (SBM)~\cite{han2014consistent,yang2009bayesian} and its mixed membership counterpart~\cite{xing2010state}. Indeed, the variant of the stochastic block model~\cite{holland1983stochastic, nowicki2001estimation} we analyze here is a special case of some of these models:\ namely, where nodes change their community membership over time, but where edges are generated independently at each time step.  As a result, the network of connections between nodes at different times is locally treelike, which makes a belief-propagation approach asymptotically optimal and allows us to compute the detectability threshold exactly.

In static networks, it has recently been shown that there exists a phase transition in the detectability of communities~\cite{decelle2011asymptotic,mossel2012reconstruction,massoulie2014community,mossel-neeman-sly-colt14} such that below the transition \textit{no algorithm can recover the true communities better than chance} (for two groups of equal size) but that efficient algorithms exist above it. Here, we generalize this result to dynamic networks, deriving a mathematically precise expression that describes where the detectability transition occurs as a function of both the strength of the communities and how quickly their membership is changing. When temporal correlations in community membership are present, we show that community detection in dynamic networks improves substantially over detection in static networks (or in a dynamic network where we cluster each graph independently). 

Finally, we give two principled and efficient algorithms for community detection in dynamic networks. Specifically, we use belief propagation (BP) to pass messages between neighbors both within a given graph and between time-adjacent graphs to integrate information over the network's history in an optimal way.  We then linearize BP to obtain a spectral algorithm, based on a dynamic version of the non-backtracking matrix~\cite{Krzakala2013,bordenave2015non}. We show experimentally that these algorithms can accurately recover the true community structure in dynamic networks all the way down to the threshold.

\section{A dynamic stochastic block model}
\label{sec:generative}
The stochastic block model (SBM) is a classic model of community structure in static networks. Here, we use a variant of the SBM in which the community labels of nodes change over time, but where edges are independent, which is a special case of several models previously introduced for community detection in dynamic networks~\cite{kim2013nonparametric,xu2014dynamic,han2014consistent,yang2009bayesian,xing2010state}. Crucially, our variant captures the important behavior of changing community labels and is analytically tractable.

Under the SBM a graph $G=(V, E)$ is generated as follows. Using a prior distribution $q_r$ over $k$ group or community labels, we assign each of the $n$ nodes $i \in V$ to a group $g_i$. We then generate the edges $E$ according to the probability specified by a $k \times k$ community interaction matrix $p$ and the group assignments $g$. In the sparse case, where $|E|=O(n)$, the resulting network is locally tree like and the number of edges between groups is Poisson distributed with parameter $c_{rs} = np_{rs}$.

In a dynamic network, we have a sequence of graphs $G(t)=(V, E(t))$ with $0 \le t \le T$, where each graph has its own group assignment vector $\left\{ g_i(t)\,|\, i\in V, t\in\{1,\ldots,T\}\right\}$. To generate each such assignment, we draw $g_i(0)$ from the prior, where each node has probability $q_r$ of being in community $1 \le r \le k$.  With probability $\eta$, each node keeps its label from one time step to the next $g_i(t)=g_i(t-1)$, and otherwise it chooses a new label $g_i(t)$ from the prior $q_r$.  Formally, the transition probability for community memberships is
\begin{equation}
  P( g(t) \,|\, g(t-1) )= \prod_{i} \left( \eta\, \delta_{g_i(t),g_i(t-1)} + (1-\eta) q_{g_i(t)}\right) \enspace,
  \label{eq:tran_prob}
\end{equation}
where $\delta_{a,b}=1$ if $a=b$ and 0 otherwise. The edges $E(t)$ are then generated independently for each $t$ according to the community interaction matrix $p$, by connecting each pair of nodes at the same time $i(t)$ and $j(t)$ with probability $p_{g_i(t),g_j(t)}$. Note that while the group assignments may change over time, the matrix $p$ remains constant. Subsequently, we use $A^{(t)}$ to denote the adjacency matrix for the graph $(V,E(t))$ at time $t$, and $D^{(t)}$ to denote the diagonal matrix of node degrees at time $t$, i.e. $D^{(t)}_{uv} = \delta_{uv} \sum_w A^{(t)}_{uw}$.

At successive times in this model, edges are correlated only through the group assignments $\{g(t)\}$. Given these, the full likelihood of a graph sequence under this dynamic SBM is 
\begin{align}
  P&( \{ E(t) \},\{ g(t) \}\,|\, p,\eta) = 
  P(\{ g(t)\})\prod_{t=0}^{T}\left(\prod_{(i,j) \in E(t)} p_{g_i(t),g_j(t)} \prod_{(i,j) \notin E(t)} 1-p_{g_i(t),g_j(t)}\right) \enspace ,
  \label{eq:tsbm_lik}
\end{align}
where $P(\{ g(t)\})=P(g(0))\prod_{t=1}^TP(g(t)\,|\,g(t-1))$.

For our subsequent analysis, we focus on the common choices of a uniform prior $q_r = 1/k$, and where $c_{rs}=np_{rs}$ has two distinct entries:\ $c_{rs}=\cin$ if $r=s$ and $c_{rs}=\cout$ if $r\ne s$. In this setting the average degree of each graph is then $ \bc = \frac{1}{k}[{\cin + (k-1) \cout}]$. We are interested in the sparse regime where $\bc = O(1)$, because most real-world networks of interest are sparse (e.g., the Facebook social network), and sparsity allows us to carry out asymptotically optimal inference. Note that the case where every group has distinct average degrees is easier than the equal-average-degree case that we consider, because distinct average degrees give prior information about group memberships.

\section{The detectability threshold in dynamic networks}
\label{sec:detectability}
The fundamental question we now consider is, under what conditions can we detect, better than chance, the correct time-evolving labeling of the latent communities in this model?

Previous work on community detection in static networks has shown that there exists a sharp threshold below which no algorithm can perform better than chance in recovering the latent community structure~\cite{decelle2011asymptotic,mossel2012reconstruction}, at least in the case $k=2$. This threshold occurs at positive values of the difference in the internal and external group connection probabilities, meaning that the community structure may still exist, but is undetectable. In terms of the SBM's parameters, this phase transition occurs at 
\begin{equation}\label{eq:static:thre}
  |\cin - \cout| = k \sqrt{\bc} \enspace .
\end{equation}

In a dynamic network where community memberships correlate across time, we will exploit these correlations to improve upon the static detectability threshold. In the worst case, when these temporal correlations are absent, i.e., $\eta=0$, we should do no worse than the static threshold. 
To facilitate our analysis, we define an extended graph structure, called a \textit{spatiotemporal graph}, in which we take $G(t)$ and add special ``temporal'' edges that connect each node $i(t)$ with its time-adjacent versions $i(t-1)$ and $i(t+1)$. Under our model, the ``spatial'' edges $E(t)$ are independent and sparse, implying that this spatiotemporal graph is locally treelike.
 
Consider a particular node $i(t)$ as $n\to\infty$ and $T\to\infty$. Moving outward in space and time, inference becomes a tree reconstruction problem, with stochastic transition matrix~$\tc$, along each spatial edge
\begin{equation}
  \tc 
= \lambda \id + (1-\lambda) \frac{J}{k} \enspace ,
\end{equation}
where $\id$ is the identity matrix, $J$ is the matrix of all $1$s, and 
\begin{equation}
  \lambda = \frac{\cin-\cout}{k\bc} \enspace . 
\end{equation}
Similarly, along each temporal edge we have a stochastic matrix
\begin{equation}
  \tau = \eta \id + (1-\eta) \frac{J}{k} \enspace .
\end{equation}
Thus, moving along a spatial or temporal edge copies a community label with probability $\lambda$ or $\eta$ respectively, and otherwise randomizes it according to the prior. That is, these edges multiply the distribution of labels by the stochastic matrices $\tc$ and $\tau$, whose eigenvalues are $\lambda$ and $\eta$, other than the trivial eigenvalue $1$ corresponding to the uniform distribution.

Since each node in the spatiotemporal graph has $\Poi(\bc)$ (Poisson-distributed random variable with mean $\bc$) spatial edges but exactly two temporal edges, the tree is generated by a two-type branching process. Each spatial edge gives rise to two temporal edges (to each of the time-adjacent versions of its end point), and each temporal edge gives rise to one temporal edge (continuing in the same direction in time), and both give rise to $\Poi(\bc)$ spatial edges.  Thus the matrix describing the expected number of children (where we multiply a column vector of populations on the left) is 
%\[
$\begin{pmatrix}
\bc & \bc \\
2 & 1 
\end{pmatrix} \, $. 
%\]
Using the results of Ref.~\cite{janson2004robust}, the detectability threshold occurs when the largest eigenvalue of matrix $\begin{pmatrix} \bc \lambda^2 & \bc \lambda^2 \\ 2 \eta^2 & \eta^2 \end{pmatrix} \,$
%\end{eqnarray}
exceeds unity, which yields
\begin{equation}
\label{eq:threshold}
\bc \lambda^2 > \frac{1-\eta^2}{1+\eta^2} \enspace . 
\end{equation}
When $\eta = 0$, i.e., when there is no temporal correlation in community assignments over time, Eq.~\eqref{eq:threshold} recovers the static detectability threshold $\bc \lambda^2 > 1$, which is equivalent to Eq.~\eqref{eq:static:thre}. On the other hand, when $\eta = 1$, i.e., when the community assignments are fixed across time, we may simply integrate the graph over $T$, making it arbitrarily dense. We then have detectability for any $\lambda > 0$, implying that any amount of community structure can be detected. At intermediate values of $\eta$, the detectability threshold falls between these two extremes.

%%%% FIGURE %%%%
 \begin{figure}[t!]
\begin{center}
\includegraphics[width=0.80\columnwidth]{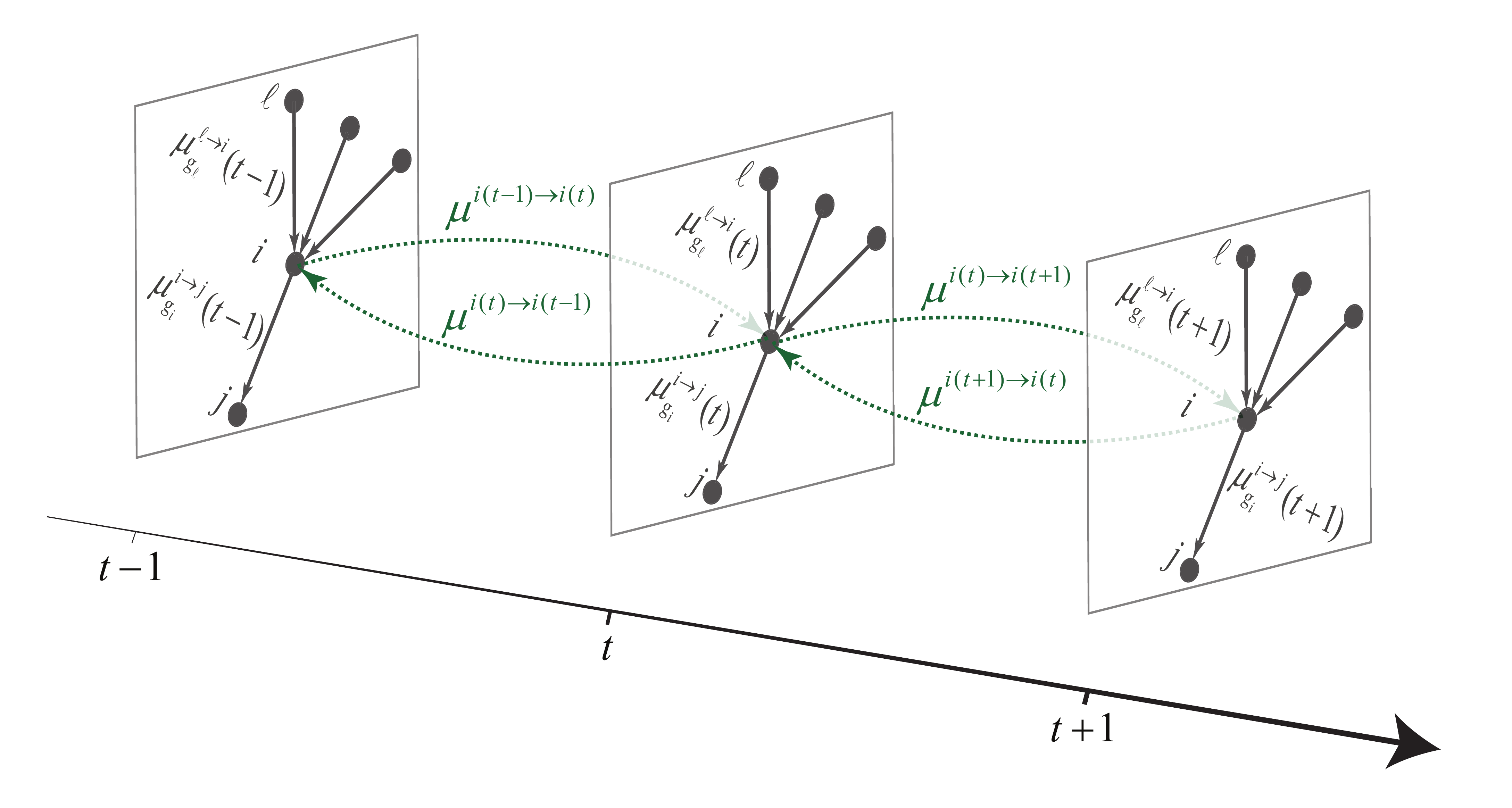}
\end{center}
\vspace{-8mm}
\caption{A schematic representation of belief propagation messages (see Eqs.\eqref{eq:spatial0} and~\eqref{eq:temporal0}) being passed along spatial and temporal edges in a spatiotemporal graph.\label{fig:msg_passing}
}
\end{figure}
%%%%%%%%%%%%%%%%

This analysis corresponds to \emph{robust reconstruction} on trees, where we are given noisy information at the leaves of a tree and we want to propagate this information to the root~\cite{janson2004robust}.  For $k=2$ groups, it is known rigorously in the static case~\cite{mossel2012reconstruction} that detecting the communities below this bound is information-theoretically impossible.  We conjecture that the same is true in the dynamic case.  For $k > 4$ groups, it has been conjectured~\cite{decelle2011asymptotic} that it is information-theoretically possible to succeed beyond the Kesten-Stigum bound, but that doing so takes exponential time.

\section{Bayesian inference of the model}
\label{sec:inference}
Given an observed graph sequence $G(t)$, we use Bayesian inference to learn the posterior distribution of latent community assignments:
\begin{equation}
	P( \{ g(t) \}\,|\,\{ E(t) \}, p,\eta) =\frac{ P( \{ E(t) \},\{ g(t) \}\,|\, p,\eta)}{\sum_{\{\gamma(t)\}}P( \{ E(t) \},\{ \gamma(t) \}| p,\eta)}.
\end{equation}

This distribution is hard to compute in general because the summation runs over an exponential number of terms. However, when the spatiotemporal graph is sparse, as generated by our model, we may make a controlled Bethe approximation (also known as belief propagation (BP) in machine learning and as the ``cavity method'' in statistical physics) that allows us to carry out Bayesian inference in an efficient and asymptotically optimal way. We now describe a BP algorithm for learning our model form data, which we then linearize to obtain a fast spectral approach, based on a dynamic version of the non-backtracking matrix. This yields two inference algorithms that perform accurately all the way down to the transition.

\subsection{Belief propagation}
\label{BP_section}
Instead of inferring the joint posterior distribution, we use belief propagation to compute posterior marginal probabilities of node labels $\{\mu_s^i(t)\}$ over time. Belief propagation assumes conditional independence of these marginals, which is exact when the graph is a tree and is a good approximation when the graph is locally-tree like, as in our spatiotemporal graph. In our setting, nodes update their current belief about marginals according to the marginals of both their spatial and temporal neighbors. That is, we define two types of messages:\ \textit{spatial} messages that pass along spatial edges and \textit{temporal} messages that pass along temporal edges. Fig.~\ref{fig:msg_passing} illustrates this message passing scheme for a spatiotemporal graph.

A spatial message $\mu^{i \to j}_r(t)$ gives the marginal probability of a node $i$ at time $t$ being in community $r$, when we consider node $j$ to be absent at time $t$. This message is computed as
\begin{align}
\mu^{i \to j}_r(t) 
&= \frac{q_r}{Z^{i \to j}(t)} 
\left(\eta \mu^{i(t-1) \to i(t)}_r + (1-\eta) \sum_u \mu^{i(t-1) \to i(t)}_u q_u \right)\nonumber\times \prod_{\substack{\ell: (i,\ell) \in E(t) \\ \ell \ne j}} \sum_s c_{rs} \mu^{\ell \to i}_s(t) 
\;\\
&\times\left(\eta \mu^{i(t+1) \to i(t)}_r + (1-\eta) \sum_u \mu^{i(t+1) \to i(t)}_u q_u \right) \times \;\prod_{\substack{\ell: (i,\ell) \notin E(t) \\ \ell \ne j}} \sum_s (1-p_{rs}) \mu^{\ell \to i}_s(t)  
 \enspace ,
\label{eq:spatial0}
\end{align}
where $Z^{i\to j}(t)$ is the normalization. The temporal message $\mu^{i(t) \to i(t+1)}_r$ (or $\mu^{i(t) \to i(t-1)}_r$) represents the marginal probability of node $i$ at time $t$ being in community $r$, when we consider node $i$ to be absent at time $t+1$ (or at $t-1$) and has a similar form:
\begin{align}
\mu^{i(t) \to i(t\pm 1)}_r 
&= \frac{q_r}{Z^{i(t) \to i(t\pm 1)}} 
\left(\eta \mu^{i(t\mp 1) \to i(t)}_r + (1-\eta) \sum_u \mu^{i(t\mp 1) \to i(t)}_u q_u \right) \nonumber \\
&\times \prod_{\ell: (i,\ell) \in E(t)} \sum_s c_{rs} \mu^{\ell \to i}_s(t) 
\;\times \;\prod_{\ell: (i,\ell) \notin E(t)} \sum_s (1-p_{rs}) \mu^{\ell \to i}_s(t) 
 \enspace .
\label{eq:temporal0}
\end{align}

When $t=0$ or $t=T$, we remove the term corresponding to the temporal edge coming from outside the domain of $t$. Furthermore, following past work on BP for the static SBM~\cite{decelle2011asymptotic,aicher2014wsbm}, we exploit these networks' sparsity to reduce the computational complexity of the spatial updates at the cost of introducing $o(\frac{1}{n})$ corrections in sparse graphs. Specifically, we let $\mu^{\ell \to i}_s(t)$ be the same for all of $\ell$'s non-neighbors $i$. We then model the effects of all such non-edges as an adaptive external field on each node, which depends on the current estimated marginals $\mu^\ell_s(t)$. That is, we let
\mbox{$  \prod_{\ell} \sum_s (1-p_{rs}) \mu^\ell_s(t) \approx \e^{-h_r(t)}$} ,
where 
\mbox{$   h_r(t) = \frac{1}{n} \sum_s c_{rs} \sum_\ell \mu^\ell_s(t)$} ,
which has the effect of preventing belief propagation from putting all the nodes at a given time into the same community. The adaptive fields only need to be updated after each BP iteration. This approximation yields a significant improvement in efficiency, reducing the computational complexity to be proportional to total number of edges in the spatiotemporal graph $\bc nT$, rather than $n^{2}T$.

Once the BP messages converge, we compute the marginal probability $\mu^i_r(t)$ that node $i$ belongs $r$ at time $t$.  This is identical to~\eqref{eq:spatial0} and~\eqref{eq:temporal0}, except that we take all incoming edges into account. We then obtain a partition by marginalization, which assigns each node to its most-likely group:
\begin{equation}
	\hat g_i(t)=\text{argmax}_{r}\mu^{i}_r(t) \enspace .
\end{equation}
It is well known in Bayesian inference~\cite{iba1999nishimori} that if the marginals are exact, then the marginalized partition is the optimal estimator of the latent community labels. Because spatiotemporal graphs under our model are sparse, we know that with $n\to\infty$, the marginals given by BP are asymptotically correct. Thus, our BP algorithm succeeds all the way down to the detectability threshold given by Eq.~\eqref{eq:threshold}, and gives an asymptotically optimal partition in terms of accuracy.

\subsection{Spectral clustering}
\label{sec:spectral}
The BP equations described above can be linearized to obtain a fast spectral approach for detecting community structure in dynamic networks.  It is easy to verify that in our setting, when $q_r=1/k$, the average degree in each group is $\bc$. This implies that BP equations will always have a solution
\begin{equation}
	\mu_r^{i(t)\to j(t)}=\mu_s^{i(t\pm 1)\to i(t)}=\frac{1}{k} \enspace ,
\end{equation}
which we call a \textit{factorized fixed point}. This fixed point only reflects the permutation symmetry in the system, and could be unstable due to random perturbations. If we use the correct parameters in BP equations, i.e., the same parameter used to generate the observed network, then in the language of physics we would say that system is in the Nishimori line~\cite{iba1999nishimori}. That is, if the BP messages deviate from the factorized solution, then they are correlated with the latent community labels and we say that there is no spin glass phase in system~\cite{iba1999nishimori}. This allows us to simplify the BP equations by studying how the messages deviate from the factorized solution, which results in a linearized version of BP. In the static SBM, this linearization is equivalent to a spectral clustering algorithm using the non-backtracking matrix~\cite{Krzakala2013}.

To do this, we rewrite the BP messages $\mu_r^{i(t)\to j(t)}$ as the uniform 
fixed point $\frac{1}{k}$ plus deviations away from it. The vector of deviations is given by
\begin{align}
\Delta^{i(t)\to j(t)}=\left\{\mu_1^{i(t)\to j(t)},\mu_2^{i(t)\to j(t)},...,\mu_k^{i(t)\to j(t)}\right\}-\left\{\frac{1}{k},\frac{1}{k},...,\frac{1}{k}\right\}\enspace , \nonumber
\end{align}
and the linearized BP equations are then
\begin{eqnarray}\label{eq:lbp}
	\Delta^{i(t)\to j(t)}&=&\sum_{\ell(t)\in\partial i(t)\backslash\, j(t)}U\Delta^{\ell(t)\to i(t)}+ V\Delta^{i(t-1)\to i(t)}+ V\Delta^{i(t+1)\to i(t)}\nonumber\\
	\Delta^{i(t)\to i(t\pm 1)}&=&\sum_{\ell(t)\in\partial i(t)}U\Delta^{\ell(t)\to i(t)}+ V\Delta^{i(t\mp 1)\to i(t)} \enspace ,
\end{eqnarray}
where $\partial i(t)$ means neighbors of $i(t)$, $U$ and $V$ denote derivatives evaluated at the factorized fixed point:
\begin{eqnarray}
U_{sr}=&\left . \frac{\partial \mu_s^{i(t)\to j(t)}}{\partial \mu_r^{\ell(t)\to i(t)}}\right |_{\frac{1}{k}}=\left . \frac{\partial \mu_s^{i(t)\to i(t\pm1)}}{\partial \mu_r^{\ell(t)\to i(t)}}\right |_{\frac{1}{k}} 
\qquad
V_{sr}=\left . \frac{\partial \mu_s^{i(t)\to j(t)}}{\partial \mu_r^{i(t\pm 1)\to i(t)}}\right |_{\frac{1}{k}}=\left . \frac{\partial \mu_s^{i(t)\to i(t\pm1)}}{\partial \mu_r^{i(t\mp 1)\to i(t)}}\right |_{\frac{1}{k}} \enspace . 
\label{eq:5}
\end{eqnarray}
Solving Eq.~\eqref{eq:lbp} amounts to finding eigenvectors of the Jacobian matrix $B$ composed of derivatives of the BP messages. However, the size of the matrix $B$ is 
$\left(cn\times T+2n\times (T-1)\right )^2$, which is relatively large for an eigenvector problem. Using the non-backtracking matrix approach~\cite{Krzakala2013}, we convert this problem into a smaller eigenvector problem of size $4nT\times 4nT$ by defining
\begin{align}
\label{eq:B}
B' = \begin{pmatrix}
\lambda \Aspatial & -\lambda \id & \lambda \Aspatial & 0 \\
\lambda (\Dspatial - \id) & 0 & \lambda \Dspatial & 0 \\
\eta \Atemp & 0 & \eta \Atemp & -\eta \id \\
\eta \Dtemp & 0 & \eta (\Dtemp - \id) & 0
\end{pmatrix} \enspace ,
\end{align}
where $\id$ denotes the $nT$-dimensional identity matrix; $\Atemp$ is the adjacency matrix of temporal edges with $\Atemp_{(u,t),(v,t')} = \delta_{uv} ( \delta_{t,t'+1} + \delta_{t,t'-1} ) \,$; $\Dtemp$ is the diagonal matrix of temporal degrees with $\Dtemp_{(u,t),(u,t)} = 2$ if $0 < t < T$, and $1$ if $t=0$ or $t=T$; $\Aspatial$ is the $nT$-dimensional matrix consisting of all the spatial edges, i.e., $\Aspatial = \bigoplus_t A^{(t)}$ meaning $A_{(u,t),(v,t')} = \delta_{tt'} A^{(t)}_{uv} \, $
; $\Dspatial = \bigoplus_t D^{(t)}$ is the diagonal matrix of spatial degrees where $D_{(u,t),(u,t)} = D^{(t)}_{uu}$. 

We now obtain a spectral clustering algorithm using $B'$ in the following way:\ given a spatiotemporal graph, we construct matrix $B'$, then take vectors composed of first $n$ entries of eigenvectors associated with the largest (absolute) eigenvalues, and finally perform $k$-means clustering on matrix composed of the vectors. This yields a partition of the nodes; if desired number of clusters is two, then we simply use the sign of entries of the vector to separate nodes into two communities.

%%%%%%%%%%
\begin{figure}[t!]
\centering
\hspace{-8mm}
\begin{tabular}{cc}
\hspace{-9mm} \includegraphics[width=0.48\textwidth]{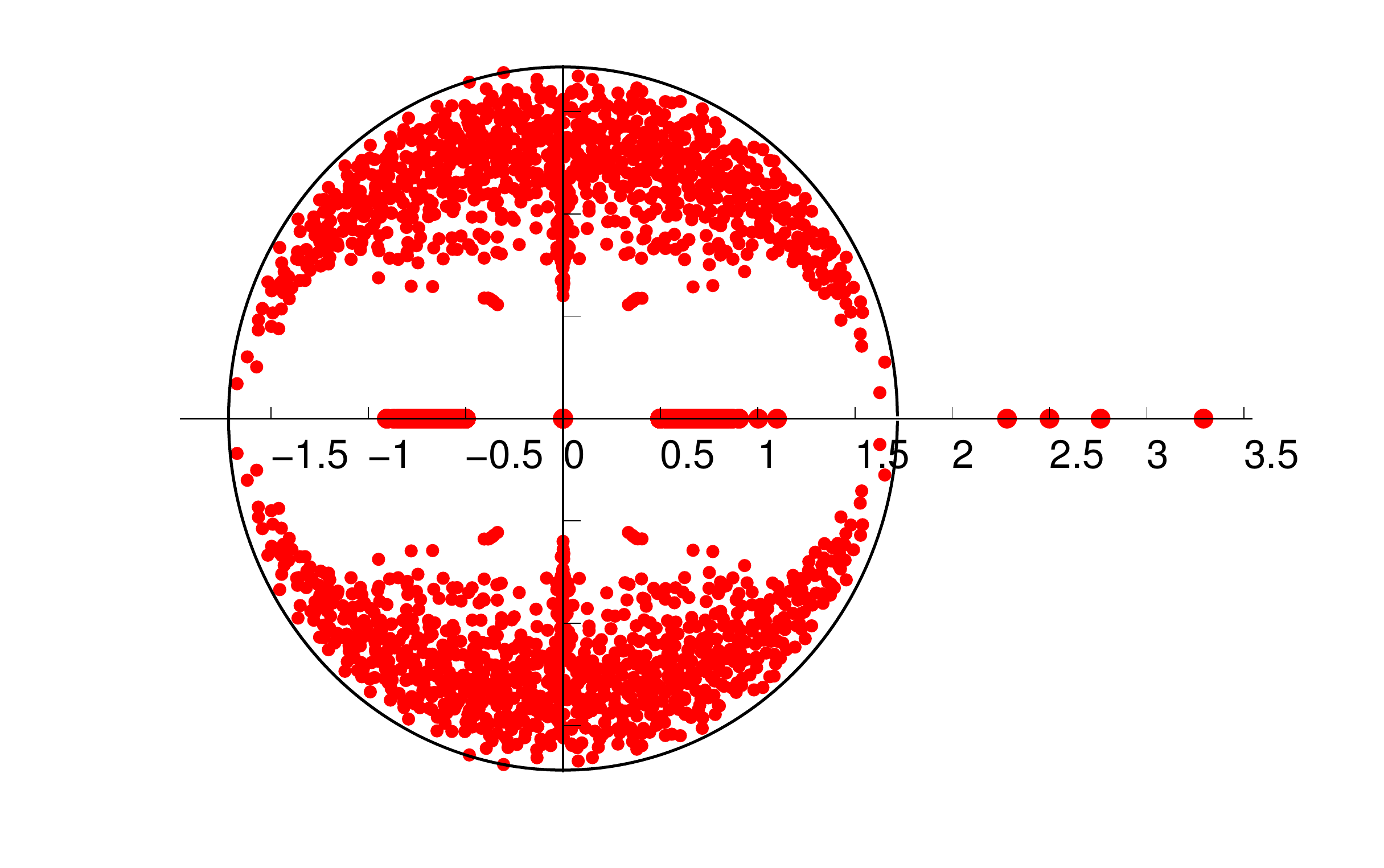} &
\hspace{-3mm} \includegraphics[width=0.60\textwidth]{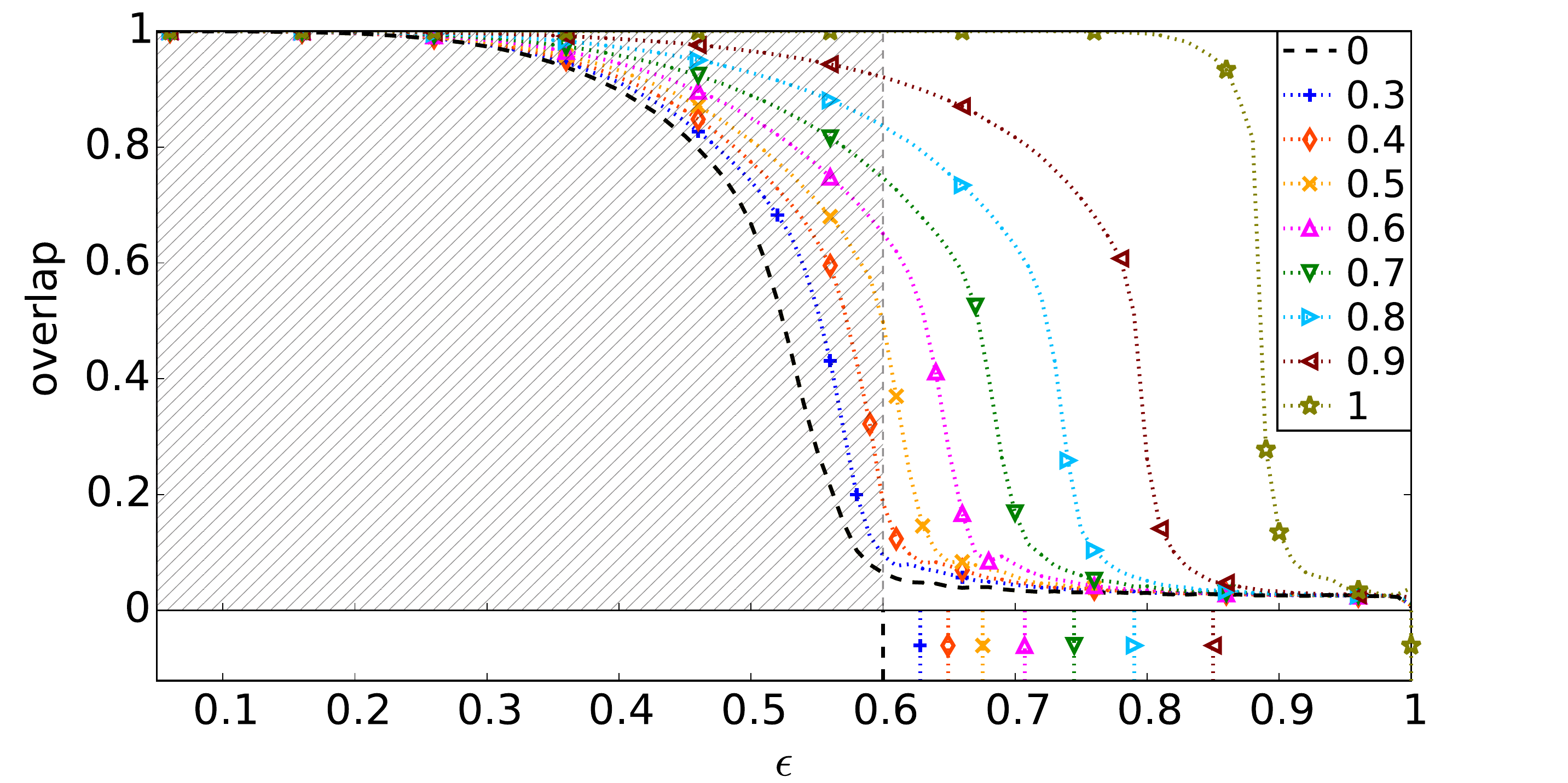}
\end{tabular}
\vspace{-7mm}
\caption{(\textit{left}) Spectrum (in the complex plane) of matrix $B'$ for a network generated by our model with $n=300, c=3, k=2$ groups and $(\epsilon,\eta)=(0.05,0.5)$. The complex eigenvalues are circumscribed by the circle. (\textit{right}) Overlap as a function of $\epsilon$ for different values of $\eta$ (given in the legend).  The detectability thresholds for each choice of $\eta$, according to Eq.~\eqref{eq:threshold}, are shown as vertical lines in lower panel, and the hatched area shows the region of detectability for static networks~\cite{decelle2011asymptotic}. Each data point is the average of 100 instances of dynamic networks from our model, with $n=512$, $T=40$, and $k=2$ groups, with average degree $c=16$.}
\label{fig:2_total}
\end{figure}
%%%%%%%%%%

From the principle of linearization, we know that real eigenvalues of the non-backtracking matrix $B'$ describe stability properties of fixed points of the BP equations, i.e., if there is a real-valued eigenvalue larger than unity, it represents a stable fixed point in the equations. Moreover, if the BP equations have a stable fixed point, then $B'$ should have a real-eigenvalue that is larger than unity, denoting a partition of the nodes that correlates with the latent community labels. Thus, our spectral clustering algorithm should work as long as BP works, implying that it also works all the way down to the detectability transition in sparse networks.

In Fig.~\ref{fig:2_total} (left) we show the spectrum of $B'$ in the complex plane for a network in the detectable regime, generated by the model. As with existing non-backtracking approaches~\cite{Krzakala2013}, most of the eigenvalues are confined to a disk, while several real eigenvalues fall outside this disk. In this example, entries of the eigenvector associated with the largest real eigenvalue have the same sign, hence the leading or ``ferromagnetic'' eigenvector does not yield information about the latent community structure.  In practice, we can perform regularizations to push such ferromagnetic eigenvectors back into bulk, thereby lifting the eigenvectors correlated with the latent community structure to the top positions. Eigenvectors associated with other real eigenvalues outside the bulk are correlated with the latent community structure. In this case, because we have two groups, we obtain the inferred partition by using the sign of entries of second real eigenvector $v_2$.

\section{Numerical verification}
\label{sec:experiments}

To verify our claims of the detectability transition in dynamic networks, and the accuracy of our algorithms, we conduct the following numerical experiment. Using our generative model of dynamic networks with community structure, we generate a number of dynamic networks for various choices of $(\epsilon,\eta)$. When $\epsilon=\cout/\cin=0$, communities are maximally strong, with every edge being located within a community, while at $\epsilon=1$, we have Erd\H{o}s-R\'enyi random graphs with no community structure. We then use our BP or spectral algorithm to infer the group assignments, assuming within each sequence that parameters $\{\eta,\epsilon,\bc\}$ are known. For each choice of $(\epsilon,\eta)$, we average our results over 100 dynamic networks with $T=40$ graphs and $n=512$ nodes (for $20,480$ nodes total), with an average degree $\bc=16$, divided into $k=2$ latent communities.

We measure the accuracy of the inferred community labels by the \textit{overlap} between the latent partition $g^*$ and the inferred one $\hat{g}$.  This is the fraction of nodes labeled correctly, maximized over all $k!$ permutations of the groups, normalized so that it is $1$ if $\hat{g} = g^*$ and $0$ if $\hat{g}$ is uniformly random. In Fig.~\ref{fig:2_total} (right) we show the overlap obtained by BP for dynamic networks as a function of $\epsilon$ for several choices of $\eta$. The detectability threshold for each $\eta$, from~\eqref{eq:threshold} is shown as vertical lines in the lower panel. When $\eta=0$, we recover the static detectability threshold given by Eq.~\eqref{eq:static:thre}. As we increase $\eta$, the phase transition occurs at increasing values of $\epsilon$, as predicted, with the largest increase occurring when $\eta=1$. 

Similar results are obtained for other choices of $n$ and $T$, with better agreement for larger networks. The slight deviation between numerical and analytic transition points observable in Fig.~\ref{fig:2_total} right is a finite-size effect, which we numerically estimated to decrease like $O(\sqrt{nT})$.

\begin{figure}
\begin{center}
\begin{tabular}{cc}
\hspace{-6mm} \includegraphics[width=0.60\textwidth]{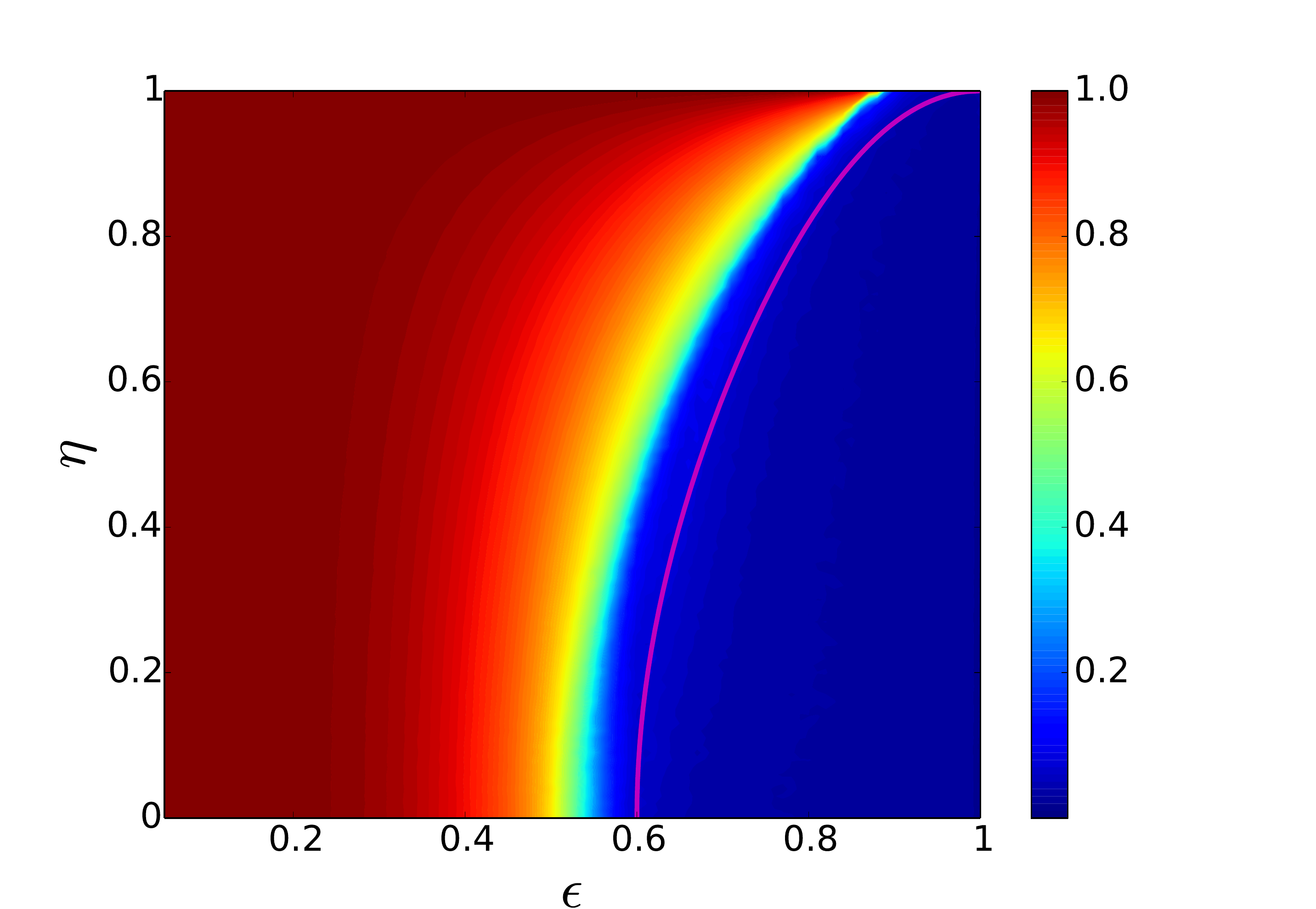} &
\hspace{-20mm} \includegraphics[width=0.60\textwidth]{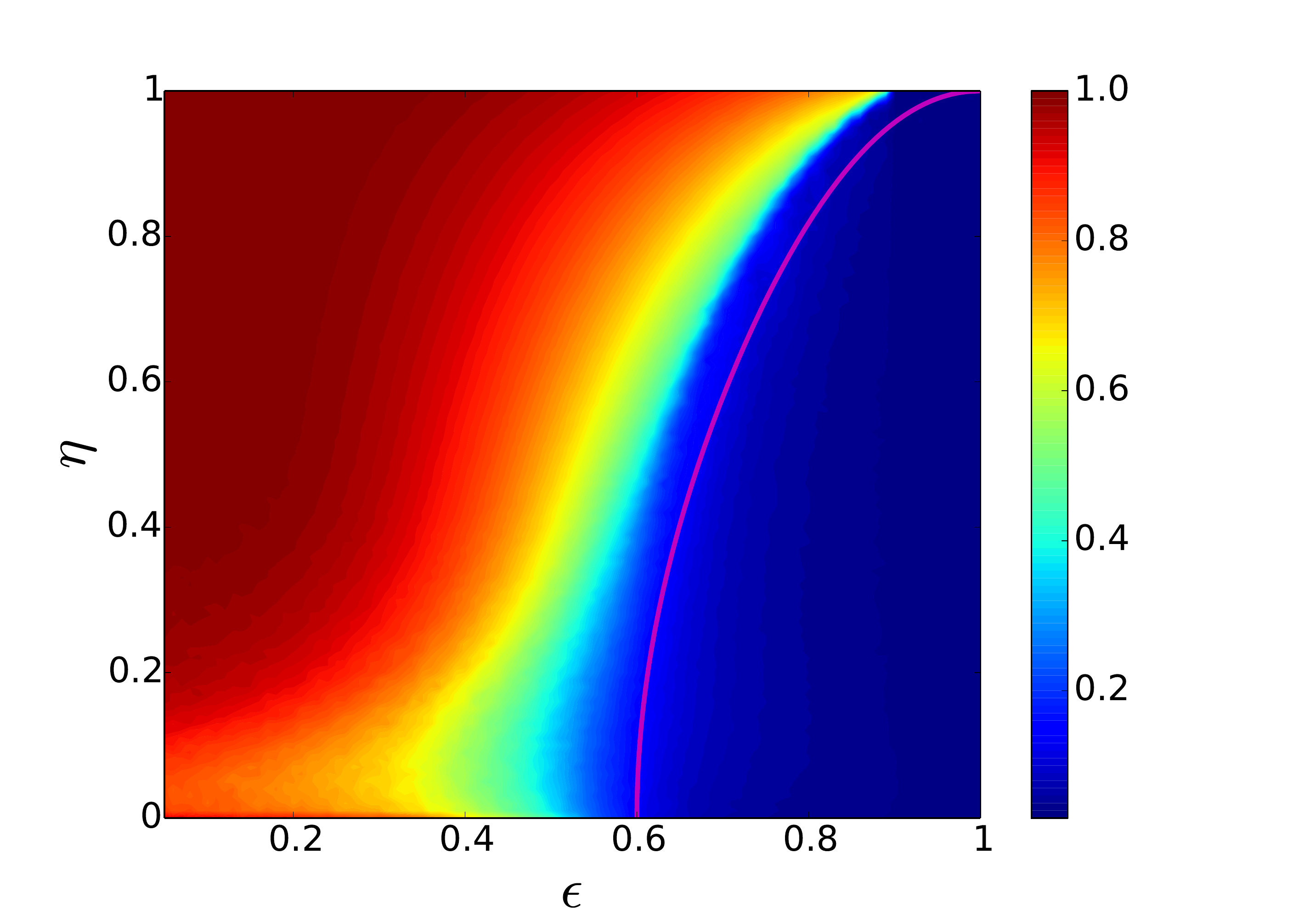}
\end{tabular}
\end{center}
\vspace{-3mm}
\caption{Heat maps showing the numerically estimated overlap for (\textit{left}) belief propagation and (\textit{right}) spectral algorithms. The detectability threshold from Eq.~\eqref{eq:threshold} is shown as a solid line. Each point shows the average over 100 instances of dynamic networks drawn from our model with $n=512,$ $T=40$, $k=2$ groups, and average degree $c=16$.
\label{fig:hmap_final_phase}
}
\end{figure}

Figure~\ref{fig:hmap_final_phase} show the overlap throughout the $(\epsilon, \eta)$-plane, using both BP and spectral algorithms, along with the line of the threshold given by Eq.~\eqref{eq:threshold}. Notably, both algorithms perform similarly:\ they have large overlap with small $\epsilon$, indicating that the learned partition is highly correlated with the latent community structure. As $\epsilon$ increases (weaker community structure), both algorithms encounter a second-order phase transition in which the overlap decreases from a finite value to zero. Separate numerical experiments indicate that the convergence time of BP diverges in the vicinity of the phase transition, which agrees with past work on the detectability threshold in static networks~\cite{decelle2011asymptotic}. We also find that at each point in ($\epsilon, \eta$)-plane, the accuracy of BP is always larger than that of the spectral algorithm, especially away from the transition, reflecting the optimality of our BP algorithm.

%%%% FIGURE %%%%
%%%%%%%%%%%%%%%%

\section{Conclusions}
\label{sec:dis}
We have derived a mathematically precise and general limit to the detectability of communities in dynamic networks. This threshold assumes a probabilistic model of community structure that is a special case of several previously developed methods to detect dynamic communities: specifically, where nodes may change their community membership over time, but where edges are generated independently at each time step. We also gave two efficient algorithms for learning latent community structure that are optimal in the sense that they succeed all the way down to the detectability threshold in dynamic networks.

A simple extension of our algorithm is to apply our BP equations to a dense network consisting of all spatial edges from all graphs projected to the time $t$, handling the message passing over time steps by using a damping factor $\tau^{|t-t'|}$. This approach extends our analysis to networks that evolve in continuous time rather than in discrete time steps.  

For larger numbers of groups, such as $k > 4$, it has been conjectured~\cite{decelle2011asymptotic} that there is a ``hard but detectable'' regime where the factorized fixed point described in Section~\ref{sec:spectral} is locally stable, but where one or more accurate fixed points exist as well.  In such a regime, community detection is information-theoretically possible, but we believe that it takes exponential time (though see~\cite{kanade-mossel-schramm} for the case where the number of groups grows with $n$).  We propose this as a direction for further work.

Other directions for future work include handling cases where the community interaction matrix $p$ may also change over time (a situation similar to change-point detection in networks~\cite{peel2015detecting}), where edges are not generated independently at each time step, or where networks have edge weights~\cite{aicher2014wsbm} or node annotations.

\begin{acknowledgements}
The authors thank Elchanan Mossel and Andrey Lokhov for helpful conversations. Financial support for this research was provided in part by Grant No.\ IIS-1452718 (AG, AC) from the National Science Foundation, Grant \#FA9550-12-1-0432 from the U.S.\ Air Force Office of Scientific Research (AFOSR) and the Defense Advanced Research Projects Agency (DARPA) (LP), and the John Templeton Foundation (PZ, CM). Author order is joint first-authorship for AG and PZ, with the remaining authors appearing alphabetically.
\end{acknowledgements}

\end{document}